\documentclass[journal]{IEEEtran}
\newcommand{\ignore}[1]{}
\usepackage[utf8]{inputenc}
\usepackage[T1]{fontenc}
\usepackage{amsmath,epsfig}
\usepackage{graphicx}
\usepackage{bm}
\usepackage{amssymb}
\usepackage{cite}
\usepackage{array}
\usepackage{extarrows}
\usepackage{url}

\usepackage{color}

\usepackage{multirow}
\usepackage{booktabs}
\usepackage{blkarray}
\usepackage{multirow}
\usepackage{array}
\usepackage{color}
\usepackage{pmat}

\begin{document}
%
\title{Video Person Re-identification by Temporal Residual Learning}
\author{Ju Dai$^{*}$,
        Pingping~Zhang$^{*}$,
        Huchuan~Lu,~\IEEEmembership{Senior~Member,~IEEE,}
        and~Hongyu Wang,~\IEEEmembership{Member,~IEEE}
\thanks{
Copyright (c) 2018 IEEE. Personal use of this material is permitted. However, permission to use this material for any other purposes must  be obtained from the IEEE by sending an email to \textcolor{blue}{\underline{pubs-permissions@ieee.org}}.

All the authors are with School of Information and Communication Engineering, Faculty of Electronic Information and Electrical Engineering, Dalian University of Technology, Dalian, 116024, P.R. China.
$^{*}$J. Dai and PP. Zhang have equal contribution for this work.
The corresponding author is Prof. Huchuan Lu.
Email: daijucug@mail.dlut.edu.cn; jssxzhpp@mail.dlut.edu.cn; lhchuan@dlut.edu.cn; whyu@dlut.edu.cn.

This work is supported in part by the National Natural Science Foundation of China (NNSFC), No. 61502070, No. 61528101 and No. 61472060. PP. Zhang is currently visiting the University of Adelaide, supported by the China Scholarship Council (CSC) program.}
}
\maketitle
\markboth{Submitted to IEEE Transactions on Image Processing,~Vol.~XX, No.~XX, Feb~2018}{}
\begin{abstract}
In this paper, we propose a novel feature learning framework for video person re-identification (re-ID).
The proposed framework largely aims to exploit the adequate temporal information of video sequences and tackle the poor spatial alignment of moving pedestrians.
More specifically, for exploiting the temporal information, we design a temporal residual learning (TRL) module to simultaneously extract the generic and specific features of consecutive frames.
The TRL module is equipped with two bi-directional LSTM (BiLSTM), which are respectively responsible to describe a moving person in different aspects, providing complementary information for better feature representations.
To deal with the poor spatial alignment in video re-ID datasets, we propose a spatial-temporal transformer network (ST$^2$N) module.
Transformation parameters in the ST$^2$N module are learned by leveraging the high-level semantic information of the current frame as well as the temporal context knowledge from other frames.
The proposed ST$^2$N module with less learnable parameters allows effective person alignments under significant appearance changes.
Extensive experimental results on the large-scale MARS, PRID2011, ILIDS-VID and SDU-VID datasets demonstrate that the proposed method achieves consistently superior performance and outperforms most of the very recent state-of-the-art methods.
\end{abstract}
\IEEEpeerreviewmaketitle
\section{Introduction}
\IEEEPARstart{G}{iven} an image/tracklet taken from one camera, person re-identification (re-ID) is the process of matching the person from images/tracklets of interest in another view.
In recent years, the computer vision field has witnessed the increasing attention in person re-ID.
A large mount of person re-ID approaches have emerged, due to its widely potential applications, such as criminal spotting~\cite{reid}, pedestrian searching~\cite{multi-search} and cross-camera tracking~\cite{Multi-track}.
%

Currently, the prominent progresses of person re-ID are achieved in the static image setting, which only uses the single images and spatial information.
Most of existing works in this setting focus on designing robust feature representation~\cite{ ELF-ECCV08,SDALF-CVPR10,eBiCov-BMVC12, ColorInv-PAMI13,gBiCov-IVC14,SCNCD-ECCV14,color-GOG,color-LOMO,CSPL}, learning discriminative distance metric~\cite{metric-RDC,metric-PCCA, metric-KISSME,metric-LADF,metric-LFDA,color-LOMO, metric-me, metric-SSSVM, metric-DNS} or combining them under a deep convolutional neural network (CNN) based framework~\cite{DL-DeepReID,DL-DTML,DL-RDC, DL-DGD, DL-TRF, DL-PSS, DL-videolatent, DL-rank, DL-videotriplet, DL-Spindle}.
Because the video setting is more close to the practical scenario, researchers have shifted their attention to the video re-ID~\cite{ DL-videoRCN, DL-videoDRCN, dataset-MARS, DL-videoForest,DL-videoASTP,DL-videoCAR,DL-video-brnn}.
The advantages of video re-ID are located in several aspects.
First, videos are the first-hand materials captured by surveillance cameras and pedestrian tracklets can be automatically detected by existing detectors.
Second, compared to the static image re-ID, videos contain more information,~\emph{e.g.}, temporal cues, pose variations and multi-view observations.
In addition, videos can provide more training samples for a tracklet sequence, usually consisting of multiple images.
%
More importantly, the ubiquitous videos offer society far-reaching benefits in terms of security and law enforcement.
Therefore, we attempt to tackle the video setting in this paper.
\begin{figure}
\begin{center}
\begin{tabular}{c@{ }c@{ }c@{ }c@{ }c}
\includegraphics[width=0.18\linewidth,height=0.25\linewidth]{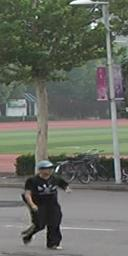}
&
\includegraphics[width=0.18\linewidth,height=0.25\linewidth]{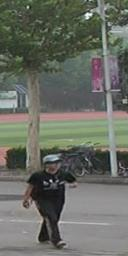}
&
\includegraphics[width=0.18\linewidth,height=0.25\linewidth]{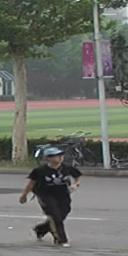}
&
\includegraphics[width=0.18\linewidth,height=0.25\linewidth]{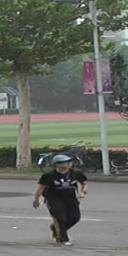}
&
\includegraphics[width=0.18\linewidth,height=0.25\linewidth]{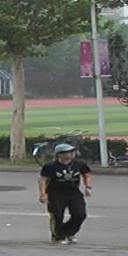}
\\
\includegraphics[width=0.18\linewidth,height=0.25\linewidth]{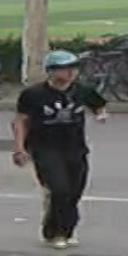}
&
\includegraphics[width=0.18\linewidth,height=0.25\linewidth]{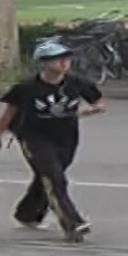}
&
\includegraphics[width=0.18\linewidth,height=0.25\linewidth]{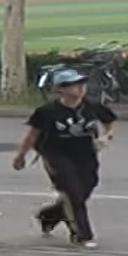}
&
\includegraphics[width=0.18\linewidth,height=0.25\linewidth]{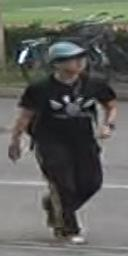}
&
\includegraphics[width=0.18\linewidth,height=0.25\linewidth]{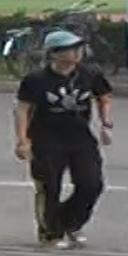}
\\
\includegraphics[width=0.18\linewidth,height=0.25\linewidth]{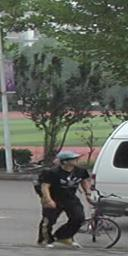}
&
\includegraphics[width=0.18\linewidth,height=0.25\linewidth]{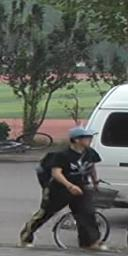}
&
\includegraphics[width=0.18\linewidth,height=0.25\linewidth]{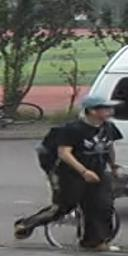}
&
\includegraphics[width=0.18\linewidth,height=0.25\linewidth]{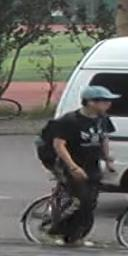}
&
\includegraphics[width=0.18\linewidth,height=0.25\linewidth]{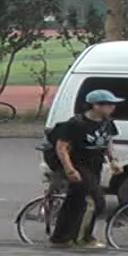}
\\
\includegraphics[width=0.18\linewidth,height=0.25\linewidth]{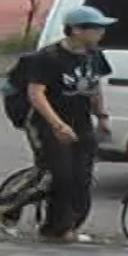}
&
\includegraphics[width=0.18\linewidth,height=0.25\linewidth]{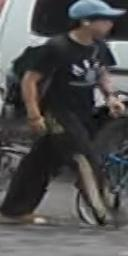}
&
\includegraphics[width=0.18\linewidth,height=0.25\linewidth]{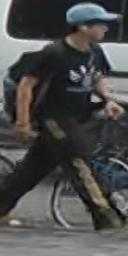}
&
\includegraphics[width=0.18\linewidth,height=0.25\linewidth]{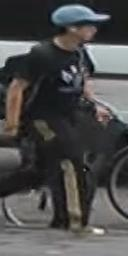}
&
\includegraphics[width=0.18\linewidth,height=0.25\linewidth]{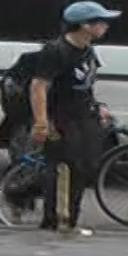}
\\\\
\end{tabular}
\end{center}
\vspace{-5mm}
\caption{Visual samples of the same person at different times under different camera views.
From left to right are the images at different times.
From top to down are the sequences under varied surveillance cameras.
The same pedestrian in video sequences can have large pose variations, out-of-focus, arbitrary scales and rather cluttered backgrounds.}
\label{video_challenges}
\vspace{-5mm}
\end{figure}

Video re-ID also faces several severe challenges.
Samples in static image re-ID are usually captured under well-controlled visual conditions or even framed by professional photographers.
However, as video acquisition is much less constrained, the image qualities of video frames tend to be rather low and pedestrians also exhibit a large range of pose variations, which can be observed from the Fig.~\ref{video_challenges}.
In particular, pedestrians in videos are usually moving, resulting in serious out-of-focus, blurring and scale variations.
Furthermore, the automatically detected pedestrian tracklets by existing pedestrian detectors may fail with cluttered backgrounds and poor alignments, which exacerbate the re-ID problems with more difficulties.

Therefore, the following issues in video re-ID need to be carefully considered:
1) how to construct appropriate person representations, so that they can effectively incorporate temporal information available in videos?
2) How to effectively harness the temporal dependency to keep better pedestrian alignments, so that sample noises and cluttered backgrounds can be alleviated and less influenced?

To deal with the first issue, recent works in video re-ID have tended to utilize the recurrent neural networks (RNNs), which take consecutive frames as inputs and adaptively incorporate temporal information~\cite{dataset-MARS, RAF-Net, DL-videoDRCN, DL-videoQAN, DL-videoForest}.
These methods first extract the frame-wise features with deep CNNs.
Then the extracted features are fed into several RNNs to capture the temporal structure information.
Finally, the average or max temporal pooling procedure is conducted on the outputs of the RNNs to aggregate the features.
However, the average pooling operation only considers the generic features of pedestrian sequences, the specific features of samples in a sequence are neglected.
While the max pooling operation concentrates on finding the local salient features, a lot of useful information may be abandoned.
A key innovation of this work is that a new technical solution to fully leverage the temporal information is provided.
Specifically, we propose a temporal residual learning (TRL) module which is equipped with two bi-directional LSTMs (BiLSTMs)~\cite{bLSTM} to simultaneously learn the generic and specific features of a video sequence.
The generic features indicate the common temporal structure of a video, which is captured by averaging the outputs of the first BiLSTM.
The specific features extracted by the second BiLSTM, characterizes the deviations and properties in specific sample frames.
The joint features can provide complementary information for person descriptions and promote the representation power of a video.
In addition, the two BiLSTMs in our framework make the generic and the specific information flow forward and backward in a flexible manner, allowing the underlying temporal information interaction to be fully exploited.

To alleviate sample noises and poor alignments appeared in videos, we propose to take advantage of the spatial transformer network (STN)~\cite{stn}.
The STN is a learnable module which can explicitly allow spatial manipulations of data within CNNs.
The STN has achieved impressive performance on fine-grained recognition~\cite{stn, rSTN}, face recognition~\cite{stn-Recursive, stn-fr} and image-level person re-ID~\cite{stn-latent, stn-PAN}.
%
The main difficulty in applying the STN to video tasks is how to ensure the spatial transformation is continuous.
In fact, consecutive frames of videos are very similar without significant changes and fluctuations.
This fact indicates that the spatial transformation in videos should also vary smoothly according to time in order to preserve the temporal continuity.
To address this problem, we extend the STN method and propose a novel spatial-temporal transformer network (ST$^2$N).
The ST$^2$N includes a recurrent structure in temporal dimension to automatically learn the optimal spatial transformation parameters in consecutive frames.
The ST$^2$N also utilizes a BiLSTM, that can leverage knowledge of consecutive frames to ensure accurate spatial alignments.

%
In summary, \textbf{our main contributions} are three-folds:
\begin{itemize}
\item
We propose a new temporal information learning method, \emph{i.e.}, the TRL module, to jointly learn the generic and specific features of a video sequence.
The complementary features can promote the representation capability of conventional RNN based temporal pooling methods.
\item
We extend the classical STN method and propose the ST$^2$N module to leverage the context knowledge of consecutive frames.
The ST$^2$N module is very useful to ensure smooth person alignments in videos.
\item
By unifying the proposed modules, our proposed video re-ID method achieves a new state-of-the-art accuracy on the large-scale MARS dataset~\cite{dataset-MARS} and impressive results on other three datasets, including the PRID2011~\cite{dataset-prid}, ILIDS-VID~\cite{dataset-ILIDS} and SDU-VID~\cite{video-STA}.
\end{itemize}

The rest of this paper is organized as follows.
In Section II, we give an overview of the video based re-ID and spatial transformer network.
Then we introduce the proposed learning approach in Section III.
In Section IV, we evaluate and analyze the proposed method by extensive experiments and comparisons with other methods.
Finally, we provide the conclusion and future work in Section V.
\begin{figure*}[t]
\begin{center}
    \includegraphics[width=1.0\linewidth,height=0.35\linewidth]{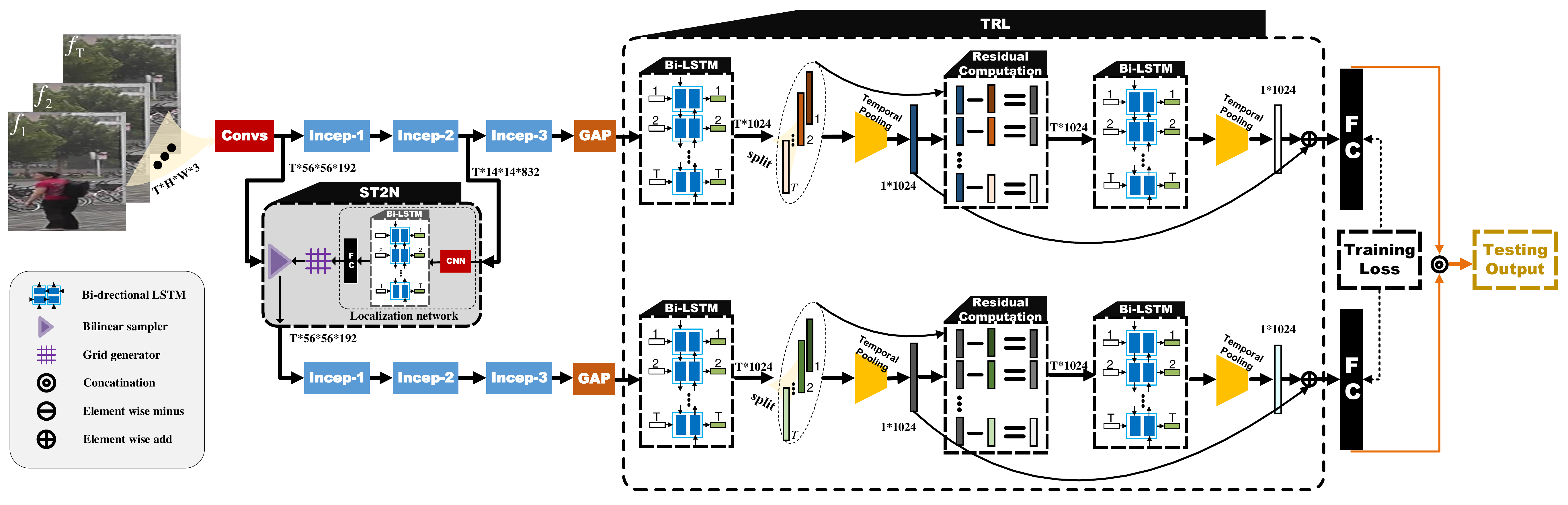}
\end{center}
\vspace{-4mm}
\caption{The overall architecture of the proposed model.
The model consists of two streams, \emph{i.e.}, the main stream (upper) and the alignment stream (lower).
The framework first takes a pedestrian sequence as input and extract deep features by using the GoogLeNet~\cite{inception-v1}.
For each stream, the first BiLSTM and Temporal Pooling unit are utilized to generate the generic features of the input sequence.
The Residual Computation part calculates the difference between the generic features and the outputs of the first BiLSTM.
The residues are further processed by the second BiLSTM and Temporal Pooling to produce the specific features.
The generic and specific features of the two streams are weighted as the final sequence representation.
In addition, the localization network in the ST$^2$N module takes the outputs of Incep-2 as inputs to predict the spatial transformation parameters for producing the aligned outputs.
}
\label{fig:fig-framework}
\vspace{-6mm}
\end{figure*}
\section{Related Work}
{\flushleft\textbf{Video based re-ID.}}
In recent years, video based re-ID has drawn increasing attention due to its numerous applications.
Existing video re-ID methods can be roughly categorized into two classes: hand-crafted feature based methods~\cite{video-STA,video-AFDA,video-DVR,video-SLDM,video-Top,video-pose,video-unsupervised} and deep learning based methods~\cite{DL-videoRCN,DL-videoDRCN, dataset-MARS, DL-videoForest,DL-videoASTP,DL-videoCAR}.

A majority of conventional algorithms develop their solutions from two aspects: extracting reliable feature representations~\cite{video-STA,video-AFDA,video-DVR,video-pose,video-unsupervised} or learning robust distance metrics~\cite{video-SLDM,video-Top}.
For instance, Wang \emph{et al.}~\cite{video-DVR} select the discriminative video fragments from noisy sequences by estimating the flow energy profile (FEP).
The fragments are represented by the HOG3D feature~\cite{HOG3D} and the average color histogram.
In~\cite{video-STA}, the periodicity of a walking person is exploited to generate spatial-temporal body-action units, which are then represented by Fisher vectors.
Cho \emph{et al.}~\cite{video-pose} conduct the multi-shot matching by efficiently estimating the target poses.
You \emph{et al}~\cite{video-Top} propose the top-push distance learning model (TPDL)~\cite{video-Top}, which integrates a top-push constrain for matching video features of persons.

Recently, with the remarkable success in various vision applications, deep learning techniques have also been applied in video re-ID.
%
In particular, Niall~\emph{et al.}~\cite{DL-videoRCN} and Wang~\emph{et al.}~\cite{DL-videoDRCN} exploit the siamese network architecture for re-ID, where the convolutional layer, the recurrent layer, and the temporal pooling layer are jointly trained to act as a feature extractor.
The difference is that the former takes advantage of RNNs to capture the temporal information based on the fixed-layer CNN features, while the later leverages the feature maps at all levels of a CNN and models the temporal information by convolutional gated recurrent units (GRUs).
In addition, since the quality of each sample cannot be guaranteed and poor images will hurt the accuracy, Liu~\emph{et al.}~\cite{DL-videoQAN} propose to learn the quality of each sample automatically.
Quality scores and features of all samples are aggregated to get the final feature representation.
Zhou~\emph{et al.}~\cite{DL-videoForest} propose the temporal attention model (TAM) to focus on discriminative frames.
The TAM is jointly learned with the spatial recurrent model (SRM) to integrate the surrounding information at different spatial locations for better similarity evaluation.
Xu~\emph{et al.}~\cite{DL-videoASTP} present an attentive spatial-temporal pooling network (ASTPN) to select key regions or frames from the sequences for the feature representation learning.

In a nutshell, existing frameworks directly aggregate the frame-level features by conducting the average, max or attention temporal pooling.
However, the average temporal pooling only summarizes the generic features, the max temporal pooling focuses on the local salient regions and the attention temporal pooling concerns the informative frames.
The specific characteristics of samples in video sequences are not explored, which may lead to a suboptimal video representation for video re-ID.
To take advantage of the temporal information and appearance speciality of pedestrians, we propose the TRL module to simultaneously extract the generic and specific features of samples within a video.
{\flushleft\textbf{Spatial transformer network.}}
To capture more spatial invariant within deep networks, Jaderberg~\emph{et al.}~\cite{stn} propose the spatial transformer module, which can automatically learn the optimal transformation parameters.
The spatial transformer module is differentiable and can be inserted into existing convolutional architectures, giving neural networks the ability to actively transform the spatial feature maps.
The resulting network improves the feature invariance to translation, scale, rotation and more generic warping.
Hence, the STN has been widely applied in fine-grained image recognition, \emph{e.g.,} face recognition~\cite{stn-Recursive, stn-fr} and static image re-ID~\cite{stn-latent, stn-PAN}.
In~\cite{rSTN}, S{\o}nderby \emph{et al.} extend the STN with RNNs in the spatial dimension for digit sequence recognition within a image.
Our proposed method shares the similar idea with~\cite{rSTN}, but we infuse the RNNs into the STN in the temporal dimension.
The resulting ST$^2$N allows the learned transformation parameters to be smooth changes within consecutive frames.
Although efforts in~\cite{stn-latent, stn-PAN} also take advantage of the STN to perform the pedestrian alignment, our model differers them in two aspects.
First, their methods are developed for static image re-ID, while our proposed module is designed to address video setting.
Second, there is no need to consider the temporal context information for static image re-ID in~\cite{stn-latent, stn-PAN}.
For the video setting, we need to control the transformation parameters in temporal dimension.
%
\section{The Proposed Approach}
In this section, we first describe the overall architecture of our framework.
Then we give the details of spatial-temporal transformer network module.
After that, we elaborate the temporal residual learning in detail.
At last, we describe the inference process of our proposed model.
\subsection{Overview of Network Architecture}
The overall architecture of our proposed model is illustrated in Fig.\ref{fig:fig-framework}.
The GoogleNet~\cite{inception-v1} is selected as the base network, however, other convolutional networks are also feasible.
The model consists of two streams: the main stream (upper) and the alignment stream (lower), which process the original input sequences and the aligned sequences, respectively.
The two streams have the same architecture and share the same front-end (Convs in Fig.\ref{fig:fig-framework}).
From the Inception-1 layers to the end, the two streams do not share parameters.
The main reason of this design is that the main stream and the alignment stream focus on different features of the sequences.
Making the parameters of two streams unshare can ensure the adaptiveness.

In detail, each frame of a pedestrian sequence is first fed into the GoogLeNet to extract multi-level convolutional features.
%
%
As pointed out in~\cite{stn-PAN}, feature maps in high-level layers can encode the attention region and semantic cues.
Inspired by this fact, we utilize the high-level feature maps (the outputs of the Inception-2 layers in the main stream, size = $T\times14\times14\times832$) as inputs of the ST$^2$N module.
The ST$^2$N predicts the transformation parameters by leveraging the spatial information of the current frame and the temporal context knowledge from consecutive frames.
The grid generator of the ST$^2$N uses the predicted transformation parameters to construct a sampling grid.
The sampling grid is a set of points, where the low-level feature maps (the outputs of Convs, $T\times56\times56\times192$) should be sampled to produce the aligned outputs.
The bilinear sampler takes the outputs of Convs and the sampling grid as inputs to produce the transformed feature maps ($T\times56\times56\times192$).
Then, we extract the feature vectors of video frames on the original feature maps and the transformed feature maps via the global average pooling (GAP).
We name the extracted features as the original sequence descriptors (OSD) and the aligned sequence descriptors (ASD).
Either the OSD or the ASD is sent into the following TRL module to get the video features.

We take the aligned pedestrian sequence as an example to demonstrate the processing procedure of the TRL module.
%
%
First, we feed the ASD to the first BiLSTM of the TRL module.
The outputs of the first BiLSTM are aggregated by a temporal poling unit.
The aggregated results act as the generic features.
Then, the differences between the generic features and the outputs of the first BiLSTM are formulated as the specific residuals.
%
%
The specific residuals are fed into the second BiLSTM, the outputs of which are summarized by the second temporal pooling.
The aggregated results are the specific features of the aligned sequence.
The final representations of the aligned sequences are obtained by fusing the generic features and the specific features.
%
\subsection{Spatial-Temporal Transformer Network Module}
%
The proposed ST$^2$N module (shown in Fig.~\ref{fig:fig-framework}) consists of three parts: the localization network, the grid generator and the bilinear sampler.
The localization network includes a shallow CNN to aggregate high-level features, a BiLSTM to capture temporal context information and a fully connected layer to predict the transformation parameters.
In detail, for the shallow CNN, we use the structure of $Conv(k=1\times1,s=1,n=512)+BN+ReLU+Max Pooling(k=2\times2,s=2)+Conv(k=1\times1,s=1,n=512)+BN+ReLU+GAP$.
For the BiLSTM, we follow previous works~\cite{rSTN,DL-videoDRCN}, and set the hidden unit size and dropout rate to 256 and 0.5, respectively.
The number of neural units in the fully connected layer is set to 4 for adapting the transformation parameters.

Formally, we describe the proposed spatial-temporal transformer process as follows.
Given an input frame $\bold{I}_t$ at time $t$, we first extract its high-level feature maps $\bold{X}_t$ (the outputs of Inception-2 layers in the main stream).
Then the shallow CNN aggregates the feature maps as
\begin{equation}
\bold{c}_t= \phi_{cnn}(\bold{X}_t),
\end{equation}
where $\bold{c}_t$ is the aggregated feature and $\phi_{cnn}(\cdot)$ denotes the shallow CNN of the localization network.
By taking the aggregated feature $\bold{c}_t$ and the bi-directional contexts ($\bold{h}_{t-1}^{fw}$, $\bold{h}_{t+1}^{bw}$) of previous frames and next frames, the BiLSTM captures the bi-directional temporal context of the current frame as
\begin{equation}
\bold{h}_t^{fw}= \phi^{fw}_{lstm}(\bold{c}_t, \bold{h}_{t-1}^{fw})
\end{equation}
\vspace{-5mm}
\begin{equation}
\bold{h}_t^{bw}= \phi^{bw}_{lstm}(\bold{c}_t, \bold{h}_{t+1}^{bw})
\end{equation}
where $\phi^{fw}_{lstm}(\cdot)$ and $\phi^{bw}_{lstm}(\cdot)$ are the forward and backward processes of the BiLSTM.
After that, we use the fully connected layer to predict the transformation parameters $\bold{\theta}_t$, which is conditioned on the temporal context information ($\bold{h}_{t}^{fw}$, $\bold{h}_{t}^{bw}$).
The transformation parameters can be expressed as
\begin{equation}
\bold{\theta}_t= \phi_{fc}(\bold{h}_t^{fw}, \bold{h}_t^{bw}).
\end{equation}
In addition, because pedestrians almost stand vertically in videos, the scale and translation transformations are enough to ensure person alignments.
Thus, we only consider the scale and and translation transformers, and perform the spatial transformation by the following affine matrix:
\begin{equation}
\bold{\theta}_t= \left[{\begin{array}{*{20}{c}}
  {s_t^x \quad 0 \quad \tau_t^x} \\
  { 0 \quad s_t^y \quad \tau_t^x} \\
\end{array}}\right],
\end{equation}
where $s_t^x$, $s_t^y$, $\tau_t^x$ and $\tau_t^y$ are the scale and and translation parameters along the width and height directions, respectively.

Similar to the STN~\cite{stn}, in our ST$^2$N we also utilize the predicted transformation parameters $\bold{\theta}_t$ to construct a sampling grid $\bold{G}_t$.
The bilinear sampler takes the input feature maps $\bold{Y}_t$ and the sampling grid $\bold{G}_t$ as inputs, and produce the spatial transformed features $\bold{\hat{Y}}_t$, which is sampled from the inputs $\bold{Y}_t$ at the grid points.
%
%
Formally, we extract the spatial transformed features $\bold{\hat{Y}}_t$ from the input maps $\bold{Y}_t$ (the outputs of the Convs in Fig.~\ref{fig:fig-framework}) by
\begin{equation}
\bold{\hat{Y}}_t= ST(\bold{Y}_t, \bold{\theta}_t),
\end{equation}
where $ST$ is the spatial transformer.
Note that, our proposed ST$^2$N model essentially searches an target region and adaptively extracts the corresponding features based on the spatial-temporal information.
$\phi_{lstm}^{fw}$ and $\phi_{lstm}^{bw}$ provide the bi-directional information for the ST$^2$N.
It can not only preserve the temporal continuity, but also allow the smooth and robust spatial transformation between consecutive frames.
We apply the spatial transformation operation on the high-level feature maps $\bold{Y}_t$ instead of the input image $\bold{I}_t$ to reduce computation efforts greatly.
Based on the ST$^2$N, we extract the features of the original pedestrian sequence and the aligned pedestrian sequence.
The resulting OSD and ASD will be sent into the temporal residual learning module to extract complementary features to represent pedestrian sequences.
\subsection{Temporal Residual Learning}
The relationship of consecutive frames provides important hints for person recognition.
Existing re-ID methods mainly introduce the forward RNNs and temporal pooling methods to capture temporal structure information.
However, the plain forward RNNs have several obvious drawbacks that may hamper the performance of video re-ID.
First, the forward RNNs merely summarize the information of previous inputs.
Thus, the outputs of forward RNNs may be biased towards the later time-steps, making the later frames more dominant than the earlier ones.
Second, the average or max temporal pooling used with forward RNNs only focuses on generic features or local salient parts of a sequence, which may neglect the valuable and specific information of samples in videos.

To address aforementioned drawbacks, we propose the temporal residual learning (TRL) module.
The TRL module can simultaneously learn the generic and specific features of pedestrian sequences.
Specifically, the TRL module contains two BiLSTMs.
Each of the BiLSTMs is followed by an average temporal pooling layer.
The employed BiLSTM inherently has forward connections and backward connections, allowing information flow backward and forward over the temporal dimension.
Thus, it is reasonable to use the BiLSTM to extract the temporal structures of both the original sequences and the aligned sequences.
In addition, the aligned sequences can provide complimentary information for the original sequences.
Therefore, the enhanced sequence features have better representation capability and robustness.

We take the aligned video sequence as an example to demonstrate the implementation details of the TRL module.
Assume the aligned video sequence $A$ contains $T$ frames, \emph{i.e.}, $\bold{I}^A=[\bold{I}_1^A, \bold{I}_2^A, ..., \bold{I}_T^A]$, and $\bold{I}_t^A$ is the image at the $t$-th time step ($t=1,2,...,T$).
%
%
Each frame $\bold{I}_t^A$ has a corresponding ASD, $\bold{f}_t^A=f(\bold{I}_t^A)$, extracted from the alignment stream.
For the notional simplicity, we subsequently drop the superscript $A$ and consider each ASD independently.
By feeding the descriptors $\bold{f}_1, \bold{f}_2, ..., \bold{f}_T$ into the first BiLSTM, we get the bi-directional temporal features, which can be expressed as:
\begin{equation}
\bold{g}^{lstm1_{fw}}_{t}= \psi_{lstm1}^{fw}(\bold{f}_t, \bold{h}^{lstm1_{fw}}_{t-1}),
\end{equation}
\vspace{-4mm}
\begin{equation}
\bold{g}^{lstm1_{bw}}_{t}= \psi_{lstm1}^{bw}(\bold{f}_t, \bold{h}^{lstm1_{bw}}_{t+1}),
\end{equation}
where $\psi_{lstm1}^{fw}(\cdot)$ and $\psi_{lstm1}^{bw}(\cdot)$ represent the forward and backward processes of the first BiLSTM in the TRL module, respectively.
$\bold{h}^{lstm1_{fw}}_{t-1}$ is the hidden state of the forward LSTM, which contains the context information of previous frames. %
$\bold{h}^{lstm1_{bw}}_{t+1}$  is the hidden state of the backward LSTM, that captures the context information of future frames.

We concatenate the bi-directional features as the temporal features of the $t$-th frame:
 \begin{align}
\bold{g}^{lstm1}_{t} = \left[\bold{g}^{lstm1_{fw}}_{t}; \bold{g}^{lstm1_{bw}}_{t}\right],
\end{align}
Then the above features are aggregated by the temporal pooling layer to extract the generic representation of a pedestrian sequence, which can be formulated as:
\begin{align}
\bold{\bar{g}}^{lstm1}_A= \frac{1}{T}\sum_t^T \bold{g}^{lstm1}_{t}.
\end{align}

Most existing methods use the $\bold{\bar{g}}^{lstm1}_A$ to perform the supervise learning for video re-ID.
However, the average temporal pooling procedure on the overall features only concerns the generic features of a video sequence, the specific features are not explicitly highlighted.
Therefore, we introduce another BiLSTM to model the specific features.
The input to the second BiLSTM is the difference between the generic feature $\bold{\bar{g}}^{lstm1}$ and the bi-directional feature $\bold{g}^{lstm1}_{t}$ at the $t$-th frame.
The detailed processing procedure of the second BiLSTM can be expressed as follows:
\begin{equation}
\bold{g}^{lstm2_{fw}}_{t}= \psi_{lstm2}^{fw}(\bold{\bar{g}}^{lstm1}-\bold{g}^{lstm1}_{t}, \bold{h}^{lstm2_{fw}}_{t-1}),
\end{equation}
\vspace{-4mm}
\begin{equation}
\bold{g}^{lstm2_{bw}}_{t}= \psi_{lstm2}^{bw}(\bold{\bar{g}}^{lstm1}-\bold{g}^{lstm1}_{t}, \bold{h}^{lstm2_{bw}}_{t+1}).
\end{equation}
We note that the feature residues $\bold{\bar{g}}^{lstm1}-\bold{g}^{lstm1}_{t}$ capture the characteristics of independent features.
They contain the specific features of each sample in a sequence.
We also aggregate the specific features to obtain the characteristic representation of the video sequence by
\begin{align}
\bold{\bar{g}}^{lstm2}_A= \frac{1}{T}\sum_t^T \bold{g}^{lstm2}_{t}.
\end{align}
Therefore, the final representation of a sequence is acquired by weighting the generic and specific features:
 \begin{align}
\bold{g}^A = \alpha\bold{\bar{g}}^{lstm1}_A + (1-\alpha)\bold{\bar{g}}^{lstm2}_A
\end{align}
where $\alpha$ is a trade-off parameter to balance the importance of the generic features and the specific features.
In the following experiments, we set $\alpha=0.5$ to give the two kind of features the equal importance.

The above procedure can also be conducted on the original video sequence $\bold{I}^O=[\bold{I}_1^O, \bold{I}_2^O,..., \bold{I}_T^O]$, the resulting features can be expressed as:
 \begin{align}
\bold{g}^O = \alpha\bold{\bar{g}}^{lstm1}_O + (1-\alpha)\bold{\bar{g}}^{lstm2}_O.
\end{align}

From the above derivation, one can see that our proposed TRL forces the temporal module not only perceive the generic features of the overall sequence, but also exploit the specific features of independent frames.
The resulting representation makes full use of the temporal structure cues, which has more representation power in contrast to only using temporal pooling methods.
We note that the BiLSTMs used in the two streams do not share the weights though we use the same symbols ($lstm1$ and $lstm2$) to represent them.
\subsection{Overall Feature Representation and Inference}
In the training phase, we employ the softmax cross-entropy loss to supervise the feature learning of the main stream and the alignment stream.
In the testing phase, for a new video sequence, we follow the feature fusion strategy in~\cite{stn-PAN} to perform the identification matching.
More specifically, given the features $\bold{g}^O$ and $\bold{g}^A$, extracted from the original video sequence and the aligned video sequence, respectively.
The overall sequence representation can be expressed as:
  \begin{align}
\bold{g} = \left[\beta|{\bold{g}^O}|^2 , (1-\beta)|{\bold{g}^A}|^2\right]
\end{align}
where $|\cdot|$ denotes the L2-norm, $\beta$ is a hyperparameter and controls the importance of each term.
In this paper, we set $\beta=0.5$ to emphasize equal importance between the main stream and the alignment stream.
In the experiments, we will show that the two streams can provide complementary information and promote each other to achieve better performance.
\section{Experiments}
In this section, we thoroughly evaluate the proposed framework on several public video re-ID datasets.
We first analyze the importance of each component of the proposed model.
To validate the effectiveness and superiority of our method, we then make extensive comparisons with other state-of-the-art video re-ID methods.
At last, we conduct cross-dataset experiments to evaluate the generalization of the proposed model, and perform hyperparameter analysis to demonstrate its insensitivity.
\subsection{Datasets}
We conduct experiments on four widely used video re-ID datasets, including the MARS~\cite{dataset-MARS}, PRID 2011~\cite{dataset-prid}, ILIDS-VID~\cite{dataset-ILIDS} and SDU-VID~\cite{video-STA} datasets.
Here we give a brief description of these datasets.
{\flushleft\textbf{MARS.}}
The MARS dataset~\cite{dataset-MARS} is a large-scale video dataset for person re-ID.
It is captured from six near-synchronized cameras on the Tsinghua campus.
As an extension of Market-1501~\cite{dataset-Market} dataset, MARS contains 1,261 different identities forming a total number of 20,478 tracklets.
These tracklets are automatically collected via the DPM~\cite{DPM} detector and GMMCP~\cite{GMMCP} tracker.
Each identity is captured by at least 2 cameras and has 13.2 tracklets on average.
Among all the tracklets, there exist 3,278 distracted tracklets due to the false detection and association, making the dataset very challenging to achieve high performance.
Following previous works, in this work we also divide this dataset into 625 persons for training and the rest for testing.
%
{\flushleft\textbf{PRID 2011.}}
The PRID 2011 dataset~\cite{dataset-prid} is collected in several uncrowded outdoor scenes with relatively simple backgrounds and rare occlusions.
Image frames of this dataset are captured by two static non-overlapping surveillance cameras.
One camera view has $385$ identities while the other has $749$ identities.
This dataset has an overlap of $200$ pedestrians appeared in both views.
Each person sequence has a variable length from $5$ to $675$ frames and with an average number of $100$.
To guarantee the effective length of image sequences, we select $178$ identities with the sequence number more than $27$ frames, following the work in~\cite{dataset-ILIDS}.
%
{\flushleft\textbf{ILIDS-VID.}}
The ILIDS-VID dataset~\cite{dataset-ILIDS} is captured at an airport arrival hall under a multi-camera CCTV network.
The dataset consists of 300 distinct individuals observed in two non-overlapping views, forming 600 pedestrian sequences.
Each person has 23 to 192 images and the average number per person is 73.
It is very challenging because of the large clothing similarities among pedestrians, lighting and viewpoint variations across views as well as cluttered backgrounds and random occlusions.
{\flushleft\textbf{SDU-VID.}}
The SDU-VID dataset~\cite{video-STA} is collected by two non-overlapping camera views in outdoor scenes.
The dataset contains 600 pedestrian sequences for 300 different identities.
Each sequence has a variable length from 16 to 346 image frames and the average number is 130.
There are more image frames in each pedestrian sequence comparing with the ILIDS-VID and PRID 2011.
The SDU-VID is also a challenging dataset due to the cluttered backgrounds, occlusions and viewpoint variations.
\subsection{Experimental Setup}
{\flushleft\textbf{Evaluation settings.}}
To evaluate the performance, we employ the cumulative matching characteristics (CMC) curves~\cite{cmc} and the mean average precision (mAP)~\cite{dataset-Market} as the evaluation criterions.
For the MARS dataset, we utilize the protocol in~\cite{dataset-MARS} and divide the MARS dataset into two subsets with 625 persons for training and the rest for testing.
While for the PRID 2011, ILIDS-VID and SDU-VID datasets, we adopt the protocol in~\cite{dataset-ILIDS}.
Each of the three datasets is randomly split into the training subset and testing subset by half, with non-overlapping identities.
During testing, we regard the pedestrians in the first camera as the probes while the second camera as the galleries.
The performances of three datasets are evaluated by the CMC score for 10 trials with different train/test splits.
The average results of 10 splits are reported.
{\flushleft\textbf{Network parameter settings.}}
We perform all the experiments on the Tensorflow platform~\cite{tensorflow}.
Most of the network parameter settings can be found in Fig.~\ref{fig:fig-framework}.
The GoogLeNet~\cite{inception-v1} pre-trained on the ImageNet dataset is selected as the CNN feature extractor.
During training, we uniformly resize the input image to $224\times224\times3$.
The number of the neural units in the two temporal BiLSTMs are set to 512.
%
%
For each sequence, we randomly choose 10 consecutive frames for training and the mini-batch size is 12.
For the the convolutional and fully connected layers, the weights are initialized by the Gaussian distribution with mean 0 and variance 0.01.
While for the weights of the BiLSTMs, we initialize them by the orthogonal initialization method~\cite{henaff2016recurrent}.
To prevent the exploding gradients, we clip all gradients to make sure them lie in the interval $[-5, 5]$.
The standard Adam algorithm~\cite{Adam} is utilized to optimize the proposed framework.

{\flushleft\textbf{Training strategy.}}
Because we introduce the ST$^2$N for the spatial alignment, which deeply interacts with the feature extraction part, the training strategy of the proposed framework needs to be carefully designed.
In other words, the effective learning of complementary features should be conditioned under well pre-trained ST$^2$N.
Thus, we train the proposed framework in a stage-wise fashion:
1) pre-training the ST$^2$N module on the large-scale MARS dataset.
In detail, we pre-train the Convs and the ST$^2$N module with frame-level supervision information.
The image-level feature vectors after the GAP are utilized to predict the identities of pedestrians in a sequence.
In this stage, we use the learning rate = $2\times10^{-4}$ and max iteration = 10,000;
2) updating the whole network for complementary feature learning.
After pre-training the ST$^2$N module, we use the learned parameters to initialize the whole network.
Then the Convs, ST$^2$N and TRL module are jointly optimized under the sequence-level supervision.
In this stage, we use the decreased learning rate = $2\times10^{-5}$, and max iteration = 10,000, 8,000, 8,000, 5,000
on the MARS, PRID 2011, ILIDS-VID, SDU-VID datasets, respectively.
Both the frame-level supervision and the sequence-level supervision are optimized by the softmax cross-entropy loss.
\subsection{Ablation Analysis}
In this subsection, we perform in-depth studies on the MARS dataset to validate the contribution of each component of the proposed model.
Table.~\ref{tab-Ablation} summarizes the ablation results of our model with the same training hyperparamters described in previous subsections.
The first four rows are conducted on the original videos, and the last six rows are conducted on both the original videos and the aligned videos.
``$G$'' refers to the GoogLeNet, that extracts the convolutional features for each frame in a pedestrian sequence.
``$LSTM$'' is a single direction LSTM and is employed to extract the temporal features.
``$BiLSTM$'' used in this paper intends to use the bi-direction information to extract the complementary features.
``$STN$'' is the classical spatial transformer network, which predicts the spatial transformation parameters on single images.
While ``$ST^2N$'' is the proposed spatial-temporal transformer module to predict parameters conditioned on extra temporal context information.
The ``$g$'' and ``$s$'' stand for the generic and specific features, respectively.
\begin{table}[t]
\caption{Quantitative comparison of different settings on the MARS dataset in terms of CMC scores (\%) at rank 1, 5, 20 and mAP. The best results are shown in boldface.}
\vspace{-6mm}
\begin{center}
\scriptsize
\begin{tabular}{|c|ccc|c|}
\hline
Model Setting          & Rank-1 & Rank-5 & Rank-20  &mAP    \\
\hline
\tiny{$G+LSTM_g$} & 69.9   & 84.7  & 88.9 & 47.4\\
\hline
\tiny{$G+BiLSTM_g$} & 74.0   & 87.8  & 91.2 & 57.8\\
\tiny{$G+BiLSTM_s$} & 73.6   &  87.4 & 90.8& 56.9\\
\tiny{$G+BiLSTM_g+BiLSTM_s$}  &76.7 & 88.7 & 91.9& 61.1\\
\hline
\tiny{$G+BiLSTM_g+STN$}  & 75.9& 88.9 & 92.1& 63.5\\
\tiny{$G+BiLSTM_s+STN$}  & 74.6& 87.9 & 92.7& 62.7\\
\tiny{$G+BiLSTM_g+BiLSTM_s+STN$}   & 77.3   & 90.2  & 94.0 & 65.6\\
\hline
\tiny{$G+BiLSTM_g+ST^2N$} &78.3 &  91.0 & 94.1 &65.4\\
\tiny{$G+BiLSTM_s+ST^2N$} & 76.7 &	89.4& 93.9&	63.4\\
\tiny{$G+BiLSTM_g+BiLSTM_s+ST^2N$} & \textbf{79.3} & \textbf{91.1} & \textbf{96.0}& \textbf{66.8}\\
\hline
\end{tabular}
\vspace{-2mm}
\end{center}
\label{tab-Ablation}
\end{table}
\begin{figure}
\begin{center}
\begin{tabular}{c@{ }c@{ }c@{ }c@{ }c}
\includegraphics[width=0.18\linewidth,height=0.25\linewidth]{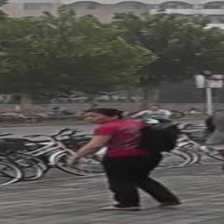}
&
\includegraphics[width=0.18\linewidth,height=0.25\linewidth]{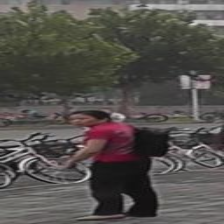}
&
\includegraphics[width=0.18\linewidth,height=0.25\linewidth]{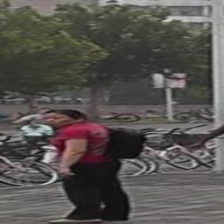}
&
\includegraphics[width=0.18\linewidth,height=0.25\linewidth]{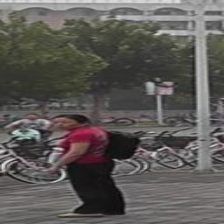}
&
\includegraphics[width=0.18\linewidth,height=0.25\linewidth]{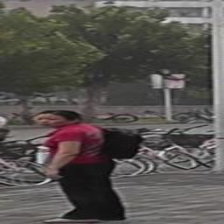}
\\
\includegraphics[width=0.18\linewidth,height=0.25\linewidth]{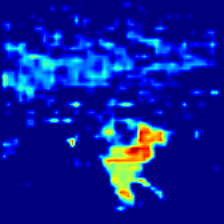}
&
\includegraphics[width=0.18\linewidth,height=0.25\linewidth]{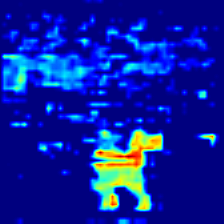}
&
\includegraphics[width=0.18\linewidth,height=0.25\linewidth]{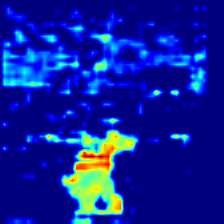}
&
\includegraphics[width=0.18\linewidth,height=0.25\linewidth]{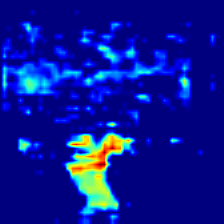}
&
\includegraphics[width=0.18\linewidth,height=0.25\linewidth]{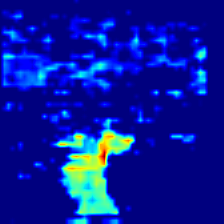}
\\
\includegraphics[width=0.18\linewidth,height=0.25\linewidth]{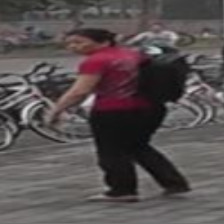}
&
\includegraphics[width=0.18\linewidth,height=0.25\linewidth]{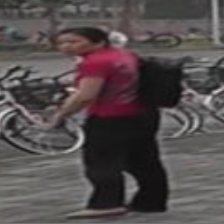}
&
\includegraphics[width=0.18\linewidth,height=0.25\linewidth]{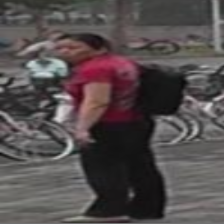}
&
\includegraphics[width=0.18\linewidth,height=0.25\linewidth]{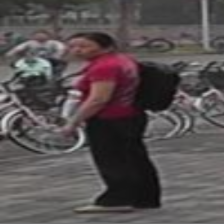}
&
\includegraphics[width=0.18\linewidth,height=0.25\linewidth]{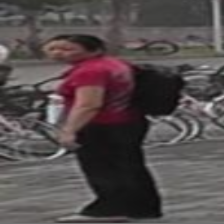}
\\
\includegraphics[width=0.18\linewidth,height=0.25\linewidth]{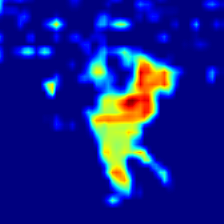}
&
\includegraphics[width=0.18\linewidth,height=0.25\linewidth]{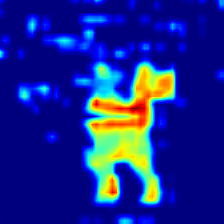}
&
\includegraphics[width=0.18\linewidth,height=0.25\linewidth]{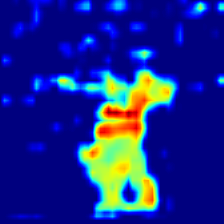}
&
\includegraphics[width=0.18\linewidth,height=0.25\linewidth]{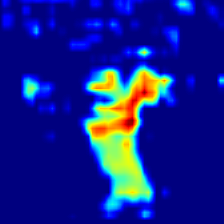}
&
\includegraphics[width=0.18\linewidth,height=0.25\linewidth]{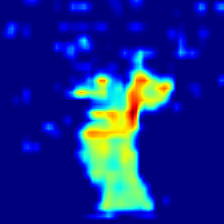}
\\
$t=1$&$t=10$&$t=20$&$t=30$&$t=40$\\
\end{tabular}
\end{center}
\vspace{-2mm}
\caption{ Visualization examples of transformed feature maps. From left to right are the frames at $t=1$, $t=10$, $t=20$, $t=30$ and $t=40$, respectively. The first two rows are the images and corresponding feature maps of original sequences. The last two rows are the images and corresponding feature maps of aligned sequences. Our ST$^2$N can partially remove cluttered backgrounds.}
\label{t-stn-maps}
\vspace{-6mm}
\end{figure}
{\flushleft\textbf{Effectiveness of the bi-directional LSTM structure.}}
From the first two rows of Table.~\ref{tab-Ablation}, one can observe that the ``$G+BiLSTM_g$'' shows a performance advantage of $4.1\%$, $3.1\%$, $2.3\%$ and $10.4\%$ in terms of the CMC scores at rank-1, 5, 20 and mAP over the ``$G+LSTM_g$''.
This indicates that compared to the conventional single directional structure, the bi-directional information flow helps the isolated frame to interact with other frames.
The BiLSTMg have better ability to capture the temporal information of pedestrian sequences.
Therefore, in following experiments we employ BiLSTMg as the base structure for extracting temporal information.
{\flushleft\textbf{Effectiveness of the STN and ST$^2$N.}}
Comparing the results of the 2-7 rows in Table.~\ref{tab-Ablation}, we can see that performing spatial alignments with spatial transformers can consistently improve the matching performance.
In particular, the models with the STN outperform the plain models with a large margin (about 5\% improvement) in term of the mAP metric.
From the results of the 8-10 rows, we can see that our proposed ST$^2$N further boosts the matching performance about 2.0\%, 1.0\%, 2.0\% and 1.5\% in terms of the CMC scores at rank-1, 5, 20 and mAP.
The results validate the effectiveness of incorporating the temporal context information for spatial transformation in practical video applications.

\begin{figure*}[t]
\begin{center}
\begin{tabular}{c@{ }c@{ }c@{ }c}
\includegraphics[width=0.24\linewidth,height=0.24\linewidth]{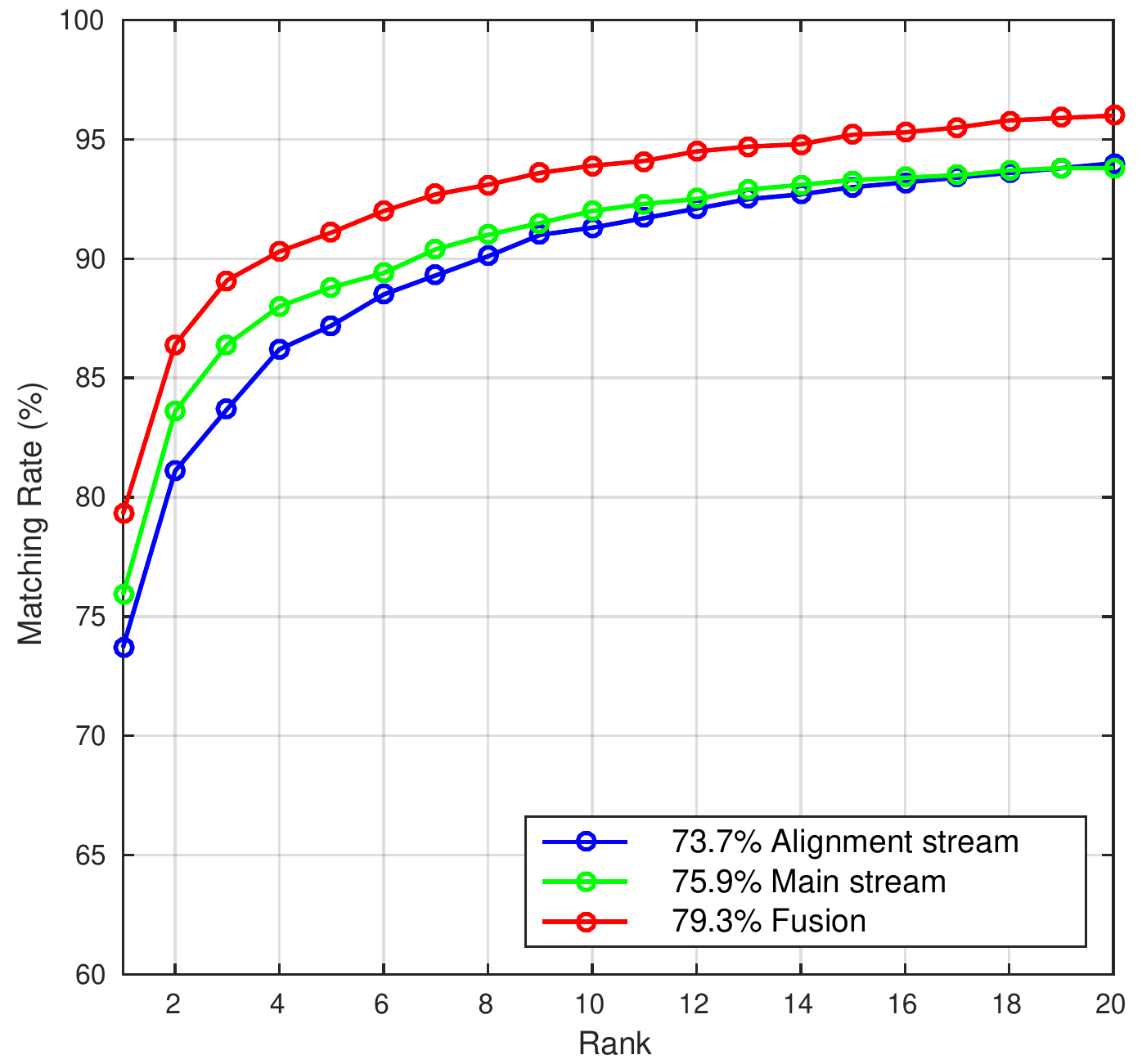}& \includegraphics[width=0.24\linewidth,height=0.24\linewidth]{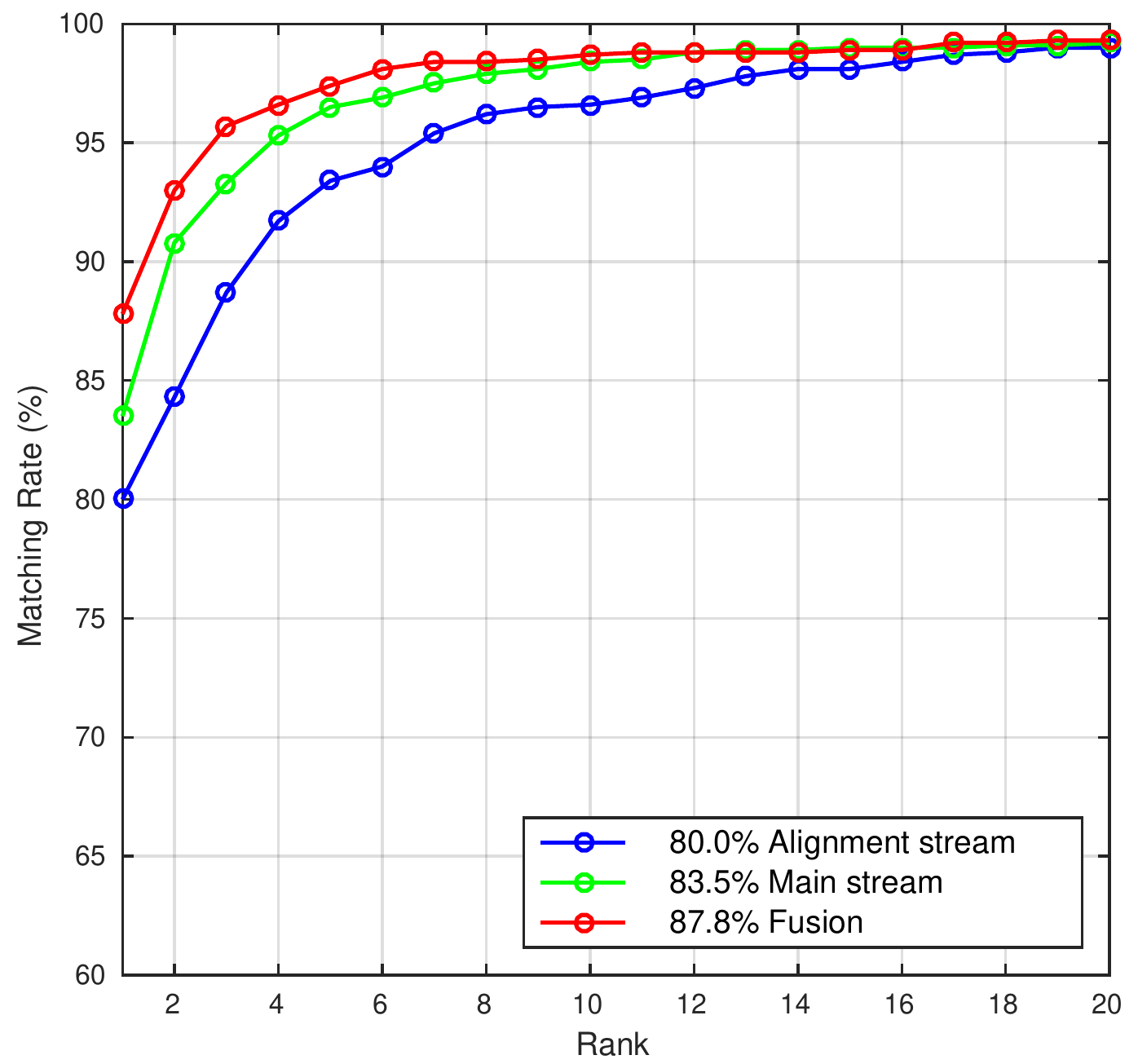}& \includegraphics[width=0.24\linewidth,height=0.24\linewidth]{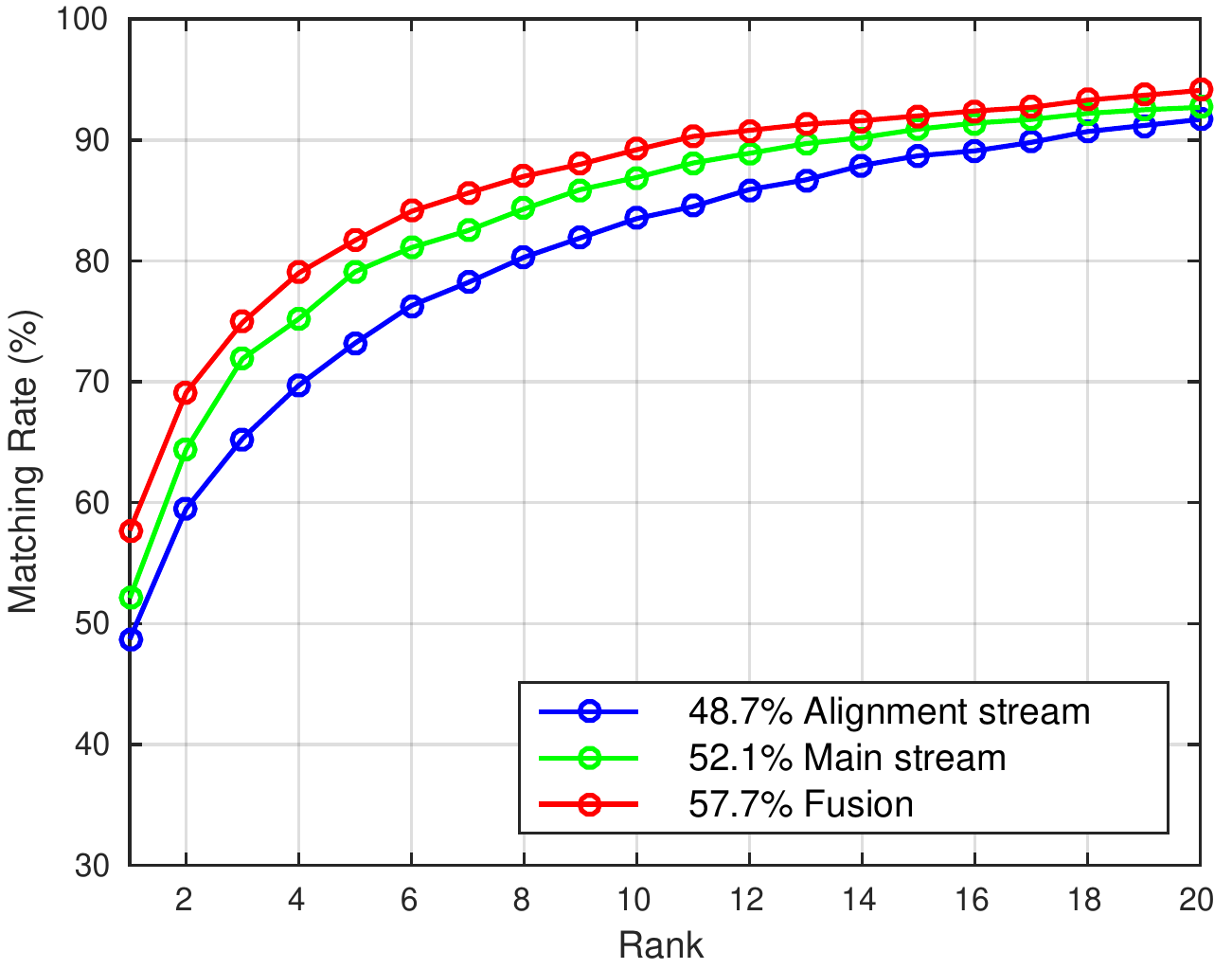}& \includegraphics[width=0.24\linewidth,height=0.24\linewidth]{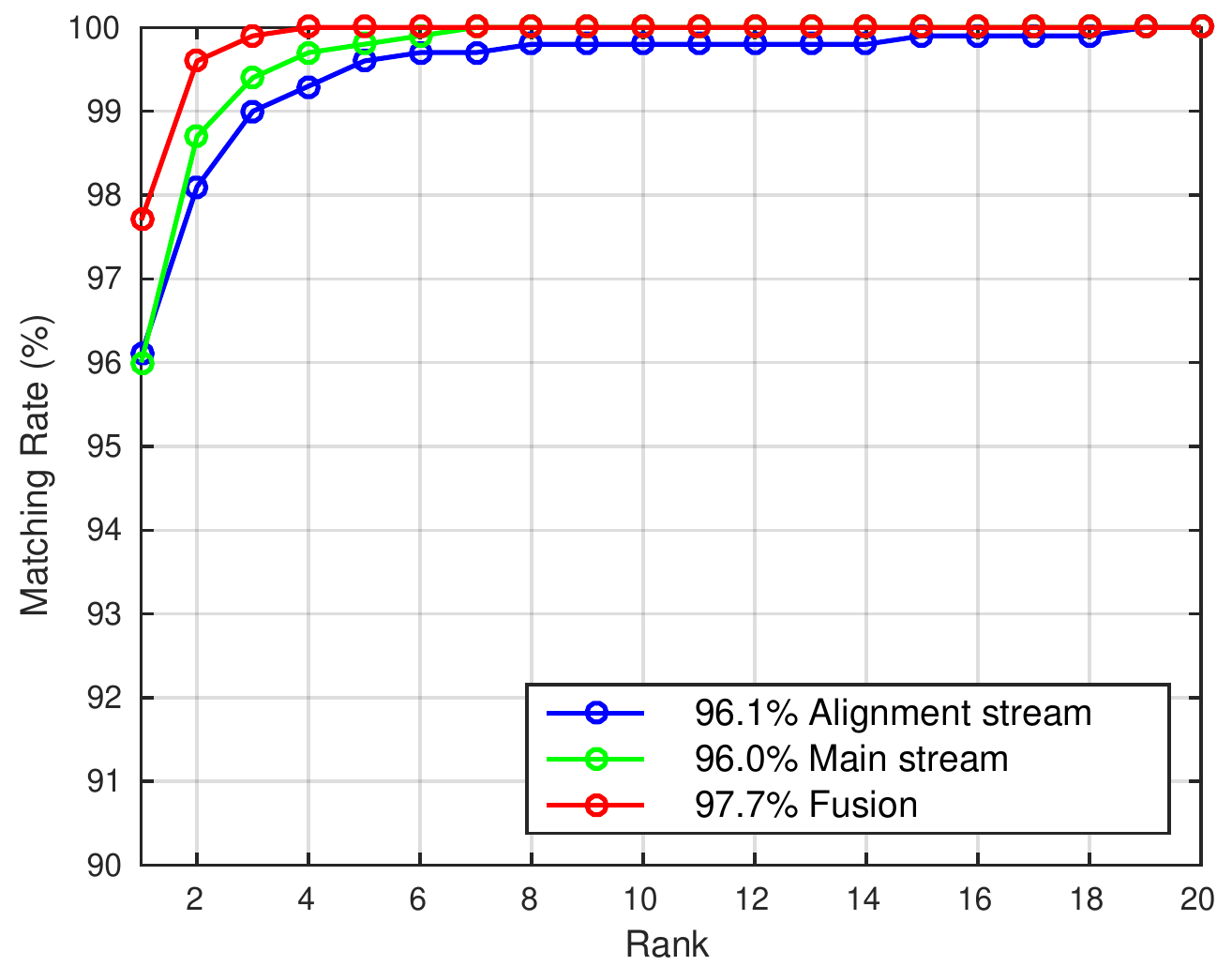}\\
(a) MARS& (b) PRID 2011& (c) ILIDS-VID& (d)SDU\\
\end{tabular}
\end{center}
\vspace{-2mm}
\caption{Results of the main stream, the alignment stream and the fusion of them on the (a) MARS, (b) PRID 2011, (c) ILIDS-VID and (d) SDU-VID datasets. The CMC scores at different rank are reported. Consistent improvements of the fusion results on all datasets can be observed.}
\label{fig:fig-compliment}
\vspace{-4mm}
\end{figure*}

In addition, to clarify the effectiveness of the ST$^2$N module, we visualize the aligned images and transformed feature maps of a randomly selected pedestrian video in Fig.~\ref{t-stn-maps}.
This figure convincingly shows that the proposed ST$^2$N module helps to align images with smooth spatial variations.
The transformed feature maps can help to improve the target attention regions and alleviate some cluttered backgrounds.
{\flushleft\textbf{Effectiveness of the generic and specific features in TRL.}}
In the TRL module, the extracted generic and specific features essentially capture different characteristics of pedestrians.
To verify the complementary behaviors, we also perform feature-level experiments.
As shown in the 2-9 rows of Table.~\ref{tab-Ablation}, only using one kind of features, the models have already achieved very impressive results.
The models with only generic features consistently outperform the ones with only specific features.
This may explains why most of previous works only use the generic features can achieve remarkable performance in video re-ID.
Though the specific features are inferior to the generic features, the combined features  achieve higher performance (about 2\% improvement) in all compared metrics.
The three group experiments convincingly demonstrate the effectiveness and superiority of the proposed TRL module.
{\flushleft\textbf{Complementarity of two streams.}}
In our proposed model, the main stream is utilized to address the original video sequences, while the aligned stream deals with the aligned video sequences.
To verify the effectiveness of each stream, we perform additional experiments on the MARS, PRID 2011, ILIDS-VID and SDU-VID datasets.
The performances of the main stream, the alignment stream and the complete model are shown in Fig.~\ref{fig:fig-compliment}.
It can be seen that the alignment stream achieves comparable performance to the main stream, and the fusion of the two streams achieves better than the individual streams on the four video datasets.
Specifically, the fusion results achieve $79.3\%$, $87.8\%$, $57.7\%$ and $97.7\%$ of CMC matching rate at rank-1 on the four datasets, which are higher than the main steam by around $3.4\%$, $4.3\%$ , $5.6\%$ and $1.7\%$, respectively.
The performance gains of the complete model are much larger than the alignment stream.
%
%
The consistent improvements indicate that the two streams can provide complementary information for each other.
When taking them together, the model achieves best results in all evaluation metrics.
\subsection{Comparison with State-of-the-art Methods}
\begin{table}[t]
\setlength{\tabcolsep}{7.9pt}
\footnotesize
\caption{Comparison with other state-of-the-art methods on Mars dataset in terms of CMC scores (\%) at rank 1, 5, 20 and mAP. The best results are shown in boldface.}
\vspace{-6mm}
\renewcommand{\arraystretch}{1.1}
\begin{center}
\begin{tabular}{|c|c|c|c|c|}
\hline
   Methods   & Rank-1 & Rank-5 & Rank-20 & mAP \\
\hline
ASTPN~\cite{DL-videoASTP} & 44.0 & 70.0 & 81.0 & - \\
CAR~\cite{DL-videoCAR} & 55.5 & 70.2 & 80.2 &-\\
CNN+XQDA~\cite{dataset-MARS} & 65.0 & 81.1 & 88.9 &45.6\\
CNN+TAM+SRM~\cite{DL-videoForest} & 70.6 & 90.0 & 97.6 & 50.7 \\
MSCAN~\cite{DL-videolatent} & 71.8 & 86.6 & 93.1 & 56.1 \\
QAN~\cite{DL-videoQAN}&73.7 &84.9& 91.6 &51.7\\
TriNet~\cite{DL-videotriplet} & 79.8 & 91.4 & - & 67.7 \\
\hline
Baseline (\tiny{$G+BiLSTM_g$}) & 74.0   & 87.8  & 91.2 & 57.8\\
Ours &79.3&91.1&96.0&66.8\\
Ours+KISSME     & 80.3 & 91.0 &95.5 &67.1\\
Ours+XQDA     & \textbf{80.5} & \textbf{91.8} & \textbf{96.0} &\textbf{69.1}\\
\hline
\end{tabular}
\vspace{-6mm}
\end{center}
\label{table:mars}
\end{table}
\begin{table*}[t]
\setlength{\tabcolsep}{7.9pt}
\footnotesize
\centering
\caption{Comparison of the proposed algorithm and other state-of-the-art methods on the PRID 2011 and ILIDS-VID datasets. The CMC scores (\%) at rank 1, 5, 20 are reported. The best three results are shown in boldface.}
\vspace{-2mm}
\renewcommand{\arraystretch}{1.1}
\begin{tabular}{l|l|c|c|c|c|c|c|c|c|c}
\hline
\multirow{2}{*}{Classes} &\ \multirow{2}{*}{Methods} & \multicolumn{3}{c|}{PRID 2011} & \multicolumn{3}{c|}{ILIDS-VID}& \multicolumn{3}{c}{SDU-VID}\\
\cline{3-11}
&\   & \multicolumn{1}{c|}{Rank-1} & \multicolumn{1}{c|}{Rank-5} & \multicolumn{1}{c|}{Rank-20}
     & \multicolumn{1}{c|}{Rank-1} & \multicolumn{1}{c|}{Rank-5} & \multicolumn{1}{c|}{Rank-20}
     & \multicolumn{1}{c|}{Rank-1} & \multicolumn{1}{c|}{Rank-5} & \multicolumn{1}{c}{Rank-20}\\
\hline
\multirow{10}{*}{Traditional}
&VR~\cite{dataset-ILIDS} & 37.6 & 63.9 & 89.4 & 34.5 & 56.7  & 77.5&-&-&-  \\
&DVR~\cite{video-DVR} &40.0& 71.7 &92.2  &39.5 &61.1  &81.8  &-&-&-\\
&DVDL~\cite{viewDic}& 40.6& 69.7 & 85.6 & 25.9 & 48.2 & 68.9 &-&-&-\\
&STFV3D~\cite{video-STA} &42.1 &71.9 &91.6 &37.0 &64.3 &86.9 &62.0&81.3&92.7\\
&AFDA~\cite{video-AFDA} &43.0 &72.7 &91.9 &37.5& 62.7 &81.8&-&-&-\\
&LFDA~\cite{metric-LFDA} &43.7 & 72.8 & 90.9& 32.9 & 68.5 &92.6 &-&-&-\\
&STFV3D+KISSME~\cite{video-STFV3D}& 64.1& 87.3 & 92.0 & 44.3 &71.7 & 91.7 &73.3 &92.7 &96.0\\
&PaMM~\cite{video-pose} &56.5 &85.7 & 97.77 &30.3 &56.3 &82.7 &-&-&-\\
&TDL~\cite{video-Top} &56.7 &80.0 &93.6 &56.3& \textbf{87.6} &\textbf{98.3} &-&-&-\\
&SIIDL~\cite{video-SLDM}& 76.7& 95.6 & 98.9 & 48.7& 81.1  & 97.3 &-&-&-\\
\hline
\hline
\multirow{12}{*}{DL-based}.
&RFA~\cite{RAF-Net} & 58.2 & 85.8 & 97.9 & 49.3 & 76.8  & 90.0 &-&-&-\\
&RNN~\cite{DL-videoRCN} &65.0 &90.0& 97.0 &50.0 &76.0 &94.0 &75.0 &86.7 &90.8\\
&RNN+OF~\cite{DL-videoRCN} & 70.0 & 90.0 & 97.0 & 58.0 & 84.0  & 96.0 &-&-&-\\
&RCN+KISSME~\cite{DL-videoDRCN} & 69.0 & 88.4  & 96.4 & 46.1 & 76.8 & 95.6 &-&-&-\\
&CNN+BRNN~\cite{DL-video-brnn} &72.8 &92.0 &97.6 &55.3 &85.0 &95.1 &85.6 &97.0 &99.6\\
&ASTPN~\cite{DL-videoASTP} & 77.0 & 95.0 & 99.0 & 62.0 & 86.0 & 98.0&-&-&-\\
&CNN+XQDA~\cite{dataset-MARS} & 77.3 & 93.5 & 99.3 & 53.0 & 81.4 & 95.1 &-&-&-\\
&CNN+SRM+TAM~\cite{DL-videoForest} & 79.4 & 94.4 & 99.3& 55.2 & 86.5  & 97.0&-&-&-\\
&CAR~\cite{DL-videoCAR} & 83.3 &93.3 &96.7 & 60.2 &85.1 &94.2&89.3 &95.3 &98.5\\
&QAN~\cite{DL-videoQAN} & \textbf{90.3} & \textbf{98.2} & \textbf{100} & \textbf{68.0} &86.8&  97.4 &-&-&-\\
\cline{2-11}
\cline{2-11}
&Baseline (\tiny{$G+BiLSTM_g$}) & 74.2 & 93.3 & 98.9 & 52.7 & 80.7 & 93.9 &91.3 &98.7&99.3\\
&Ours & 87.8 & 97.4 & 99.3 & 57.7 & 81.7 & 94.1&\textbf{97.7}& \textbf{100.0}&\textbf{100.0}\\
\hline
\end{tabular}
\label{tabel:ilids}
\end{table*}
\begin{table*}[t]
\setlength{\tabcolsep}{7.9pt}
\footnotesize
\centering
\caption{Results of cross-dataset experiments. The first collum represents the training datasets and the first row indicates the testing datasets. The CMC scores (\%) at rank-1, 5, 20 are reported.} 
\vspace{-2mm}
\renewcommand{\arraystretch}{1.1}
\begin{tabular}{l||c|c|c||c|c|c||c|c|c}
\hline
\multirow{2}{*}{Datasets} & \multicolumn{3}{c||}{PRID 2011} & \multicolumn{3}{c||}{ILIDS-VID}& \multicolumn{3}{c}{SDU-VID}\\
\cline{2-10}
     & \multicolumn{1}{c|}{Rank-1} & \multicolumn{1}{c|}{Rank-5} & \multicolumn{1}{c||}{Rank-20}
     & \multicolumn{1}{c|}{Rank-1} & \multicolumn{1}{c|}{Rank-5} & \multicolumn{1}{c||}{Rank-20}
     & \multicolumn{1}{c|}{Rank-1} & \multicolumn{1}{c|}{Rank-5} & \multicolumn{1}{c}{Rank-20}\\
\hline
MARS & 35.2 & 69.6 & 89.3 & 18.1 & 30.8 & 59.3 &89.4 & 97.6 &100.0\\
PRID 2011 & - & - & - & 8.9 & 22.8 & 48.8& 39.7& 68.4&88.7\\
ILIDS-VID &29.5 & 59.4 & 82.2 & - & - & -&45.4& 76.3&93.6\\
SDU-VID & 23.1 &54.9 & 78.6 & 7.6 & 15.9 & 41.4&-& -&-\\
\hline
\hline
\end{tabular}
\label{tabel:cross}
\vspace{-6mm}
\end{table*}
To demonstrate the superiority of our approach, we compare the proposed video re-ID model with several state-of-the-art methods on the large-scale MARS dataset and three small datasets, \emph{i.e.}, PRID 2011, ILIDS-VID and SDU-VID.
{\flushleft\textbf{Results on the large-scale dataset.}}
On the MARS dataset, we compare our method with seven state-of-the-art methods, including the ASTPN~\cite{DL-videoASTP}, CAR~\cite{DL-videoCAR}, CNN+XQDA~\cite{dataset-MARS}, CNN+TAM+SRM~\cite{DL-videoForest}, MSCAN~\cite{DL-videolatent}, QAN~\cite{DL-videoQAN} and  TriNet~\cite{DL-videotriplet}.
The detailed results are summarized in Table.~\ref{table:mars}.
As can be seen, our baseline model achieves the rank-1 accuracy of $74.0\%$ and mAP of $57.8\%$, which performs better than most of the existing state-of-the-art methods except for the TriNet~\cite{DL-videotriplet}.
Our proposed model with Euclidean distance achieves rank-1 accuracy of $79.3\%$ and mAP of $66.8\%$.
The results outperform the state-of-the-art method QAN~\cite{DL-videoQAN} by a large margin.
The improvements are $5.6\%$ and $15.1\%$ in terms of the rank-1 and mAP, respectively.
The TriNet~\cite{DL-videotriplet} is currently one of the optimal metric learning methods and our model can obtain competitive performance.
The results motivate us to employ distance metric learning to further improve the performance.
When our method combined with KISSME~\cite{metric-KISSME}, we arrive at $80.3\%$ rank-1 accuracy and $67.1\%$ mAP, with nearly $1.0\%$ and $0.3\%$ relative accuracy gains.
The proposed model with XQDA~\cite{color-LOMO} achieves the optimal results with $80.5\%$, $91.8\%$, $96.0\%$ and $69.1\%$ for the CMC scores at rank-1, 5, 20 and mAP, which outperforms all of compared methods.
These results confirm the superiority and effectiveness of the proposed model on the large-scale automatically detected MARS dataset.
{\flushleft\textbf{Results on three small datasets.}}
In addition, we also conduct experiments on three small video re-ID datasets.
On these datasets, we compare with twenty state-of-the-art methods, including VR~\cite{dataset-ILIDS}, DVR~\cite{video-DVR}, DVDL~\cite{viewDic}, STFV3D~\cite{video-STA}, AFDA~\cite{video-AFDA}, LFDA~\cite{metric-LFDA}, STFV3D+KISSME~\cite{video-STFV3D}, PaMM~\cite{video-pose}, TDL~\cite{video-Top}, SIIDL~\cite{video-SLDM}, RFA~\cite{RAF-Net}, RNN~\cite{DL-videoRCN}, RNN+OF~\cite{DL-videoRCN}, RCN+KISSME~\cite{DL-videoDRCN}, CNN+BRNN~\cite{DL-video-brnn}, ASTPN~\cite{DL-videoASTP}, CNN+XQDA~\cite{dataset-MARS}, CNN+SRM+TAM~\cite{DL-videoForest}, CAR~\cite{DL-videoCAR} and QAN~\cite{DL-videoQAN}.
Among the compared approaches, the first ten approaches use hand-crafted features and the left are deep learning based models. %
The comparison results of the CMC scores at rank-1, 5, 20 are reported in Table.~\ref{tabel:ilids}.

Results in Table.~\ref{tabel:ilids} demonstrate the significant superiority of the proposed model over existing state-of-the-art methods on the SDU-VID dataset.
Specially, our model achieves the rank-1 accuracy of $97.7\%$, and can fully distinguish different queries at rank-5.
It surpasses the previous best approach CAR~\cite{DL-videoCAR} by $8.4\%$ and $4.7\%$ in terms of rank-1 and rank-5, respectively.
For the PRID 2011 dataset, the proposed approach obtains $87.8\%$ and $97.4\%$ at rank-1 and rank-5, respectively.
Other than the QAN~\cite{DL-videoQAN} model, our model achieves the best performance in contrast to all compared methods.
On the more challenging ILIDS-VID dataset, the advantage of our method is less obvious than on the MARS and SDU-VID datasets.
But we obtain very comparable results with the CAR~\cite{DL-videoCAR} and the RNN+OF~\cite{DL-videoRCN}.
The QAN model can still achieves the best performance on the ILIDS-VID dataset.
The reason may be that training samples in the ILIDS-VID dataset are very limited, only consisting of 150 persons with 300 training sequences.
In addition, the dataset contains large variations, cluttered backgrounds and low-resolution images.
Those factors make it challenge to discriminate different pedestrians of the dataset.
The QAN~\cite{DL-videoQAN} model takes fully advantages of the labeled information by utilizing both image-level and sequence-level supervision information to train.
The results give us an evidence that it is a suitable practice to include image-level supervision when very limited labeled videos are available.
It is interesting to note that the performance margins between the hand-crafted features and the deep features are not very large, which means that hand-crafted features are befitting for small datasets.

In addition, it is not difficult to observe from Table.~\ref{tabel:ilids} that the proposed model obviously outperforms the baseline model by a large margin.
Our TRL module improves the rank-1 accuracy by $13.6\%$, $5.0\%$ and $6.3\%$ on the PRID2011, ILIDS-VID and SDU-VID datasets, respectively.
Moreover, we arrive at a new state-of-the-art performance on the SDU-VID dataset, win the second performance on the PRID 2011 datastet and achieve the competitive results on the challenge ILIDS-VID dataset.
These results clearly prove that the proposed model has obvious superiority and shows its effectiveness on small datasets as well.
\subsection{Cross-dataset Generalization}
In practice, different datasets are usually collected under different visual conditions.
%
%
Models trained with one dataset can perform badly on a different dataset, which means that dataset bias is inevitable.
Hence, it is important to design models that can generalize well to new environments.
Cross-dataset testing is an effective way to evaluate the potential ability of a system.
To better understand the generalization performance of our model, we also perform cross-dataset experiments.
More specifically, we conduct two kinds of experiments.
The first one is training on the large-scale MARS dataset and testing on the three small datasets.
The second is training on one of the three small datasets, while testing on the other two datasets.
Experimental results are listed in Table.~\ref{tabel:cross}.

The results in Table.~\ref{tabel:cross} reveal that when training on the large-scale MARS dataset, the proposed model can get better generalization performance.
In particular, we achieve $89.5\%$, $97.6\%$ and $100.0\%$ of the CMC scores at rank-1, 5 and 20 on the SDU-VID dataset.
The results are better than current state-of-the-art methods, which can be observed from Table.~\ref{tabel:ilids}.
However, the CMC scores for the PRID 2011 and the ILIDS-VID dataset is less fascinating.
The rank-1 scores for the two datasets are $30.2\%$ and $18.1\%$, respectively.
The main reason may be that sample distributions of the PRID 2011 and ILIDS-VID datasets are very different from the MARS, while the pedestrians in the SDU-VID dataset are relative simple and have similar appearances to the MARS.

When the model is trained on one of the three small datasets, the matching rates for the other two datasets decline nearly by half.
Specifically, when we train the model on the PRID 2011 dataset, we achieve $8.9\%$ rank-1 accuracy for the ILIDS-VID and $39.7\%$ for the SDU dataset.
If the challenging ILIDS-VID dataset is selected as the training dataset, our model reports $29.5\%$ and $45.4\%$ rank-1 score for the PRID 2011 and the SDU dataset, respectively.
When training on the SDU-VID, the CMC scores for the PRID 2011 and the ILIDS-VID are the lowest.
Those results demonstrate that the dataset biases do have affects on the matching rates.
Our model has certain generalization for the cross-dataset testing, but need to be further optimized to improve the generalization performance.
\subsection{Trade-off Parameter Analysis}
\begin{figure}[t]
\begin{center}
\begin{tabular}{c@{}c}
\includegraphics[width=0.85\linewidth,height=0.65\linewidth]{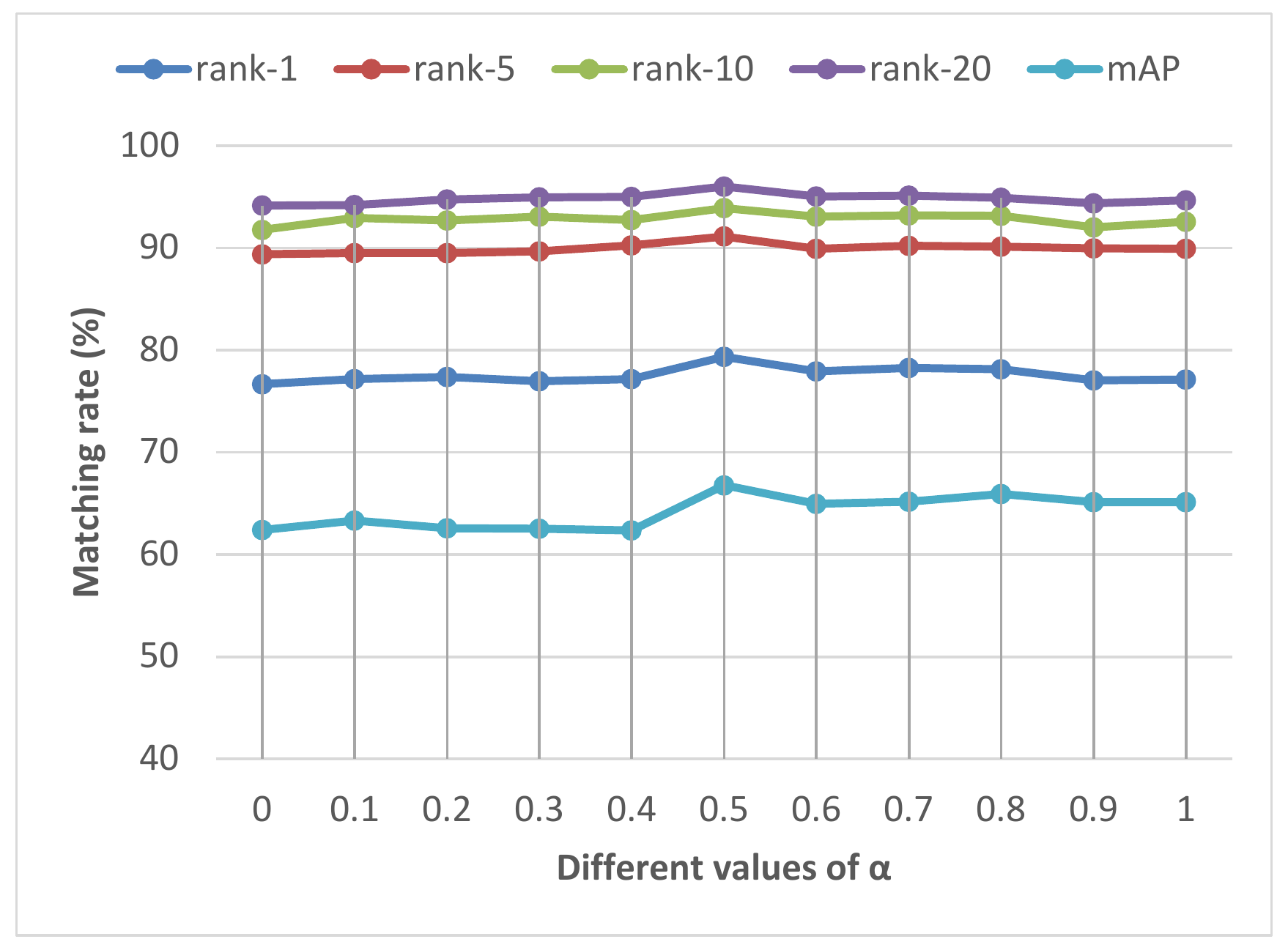}\\
(a)\\
\includegraphics[width=0.85\linewidth,height=0.65\linewidth]{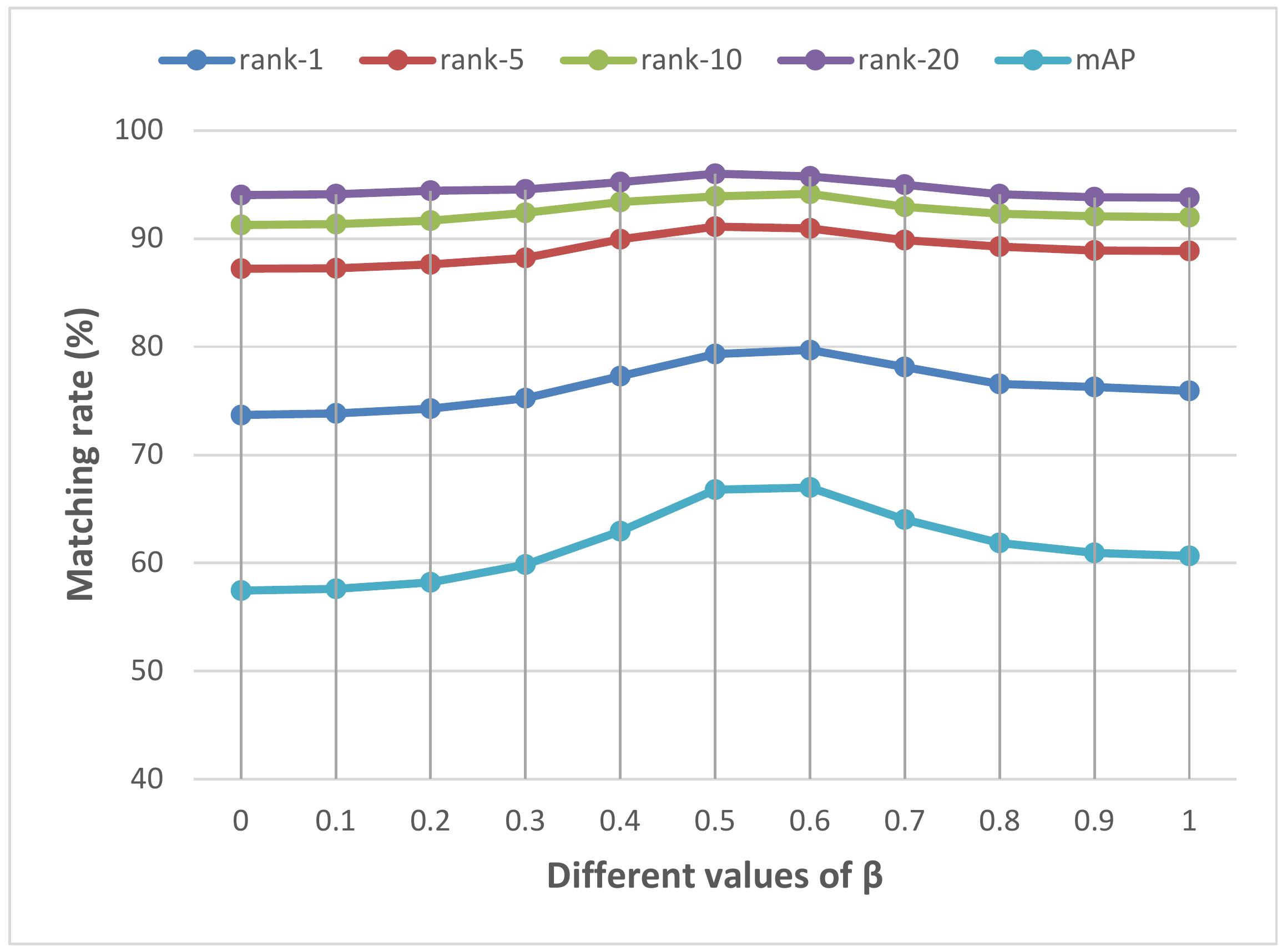}\\
(b)\\
\end{tabular}
\end{center}
\vspace{-2mm}
\caption{Parameter sensitivity analysis on the MARS dataset. (a) CMC matching rates v.s. different values of $\alpha$. (b) CMC matching rates v.s. different values of $\beta$.}
\label{fig:fig-parameter}
\vspace{-6mm}
\end{figure}
There are two trade-off parameters in our approach: $\alpha$ and $\beta$.
We conduct empirical analysis on the MARS dataset to analyze the parameter sensitivity based on the person re-ID accuracy .
When analyzing one parameter, we keep the other one fixed.
We report the CMC scores at rank-1, 5, 10, 20 as well as the mAP with $\alpha$ and $\beta$ varying in the range of $[0, 0.1, ..., 1]$.
Experimental results are illustrated in Fig.~\ref{fig:fig-parameter}.

The parameter $\alpha$ balances the importance between the generic and specific features of pedestrian sequences.
When $\alpha=0$, it means that we merely consider the specific features.
$\alpha=1$ indicates that we only utilize the generic features to describe pedestrians.
Fig.~\ref{fig:fig-parameter} (a) demonstrates that when we take both terms into consideration, we achieve better performance.
The parameter $\beta$ controls the significance of the main stream and the alignment stream.
It can be observed from Fig.~\ref{fig:fig-parameter} (b) that both the main stream and the alignment stream are important and can complement each other.
It can be also noticed that the matching rates have relative small fluctuations when $\alpha$ and $\beta$ varies.
The results indicates that the proposed algorithm does not rely on parameter tuning to obtain outstanding performance.
However, when we choose $\alpha=0.5$ and $\beta=0.5$, we can get the highest results.
Therefore, in experiments, we set $\alpha=\beta=0.5$ to emphasize equal importance for any of the balance terms in the proposed model.
\section{Conclusion}
In this work, we focus on the problems of unconstrained video person re-ID, where video sequences contain severe challenges, like out-of-focus target and cluttered backgrounds.
To address these key challenges, we propose a novel spatial-temporal transformer network (ST$^2$N), which leverages temporal information to keep smooth spatial alignments between consecutive frames.
In addition, to fully take advantage of the extra temporal information, we propose an innovative temporal residual learning (TRL) module to learn the generic as well as the specific features of pedestrian sequences.
The TRL module incorporates bi-directional LSTM structures to allow temporal information not only propagate from front to back but also in the reverse direction.
Extensive experimental results on the MARS, PRID 2011, ILIDS-VID and SDU-VID datasets demonstrate the effectiveness and superiority of the proposed method over other state-of-the-art methods.
\ifCLASSOPTIONcaptionsoff
  \newpage
\fi



\bibliographystyle{IEEEtran}
\bibliography{IEEEabrv,refs}

%
\begin{IEEEbiography}{\textbf{Ju Dai}}
received her B.S. degree and M.Sc. degree in Electronic Engineering, China University Of Geosciences (CUG), Wuhan, in 2011 and 2014, respectively. She is currently a Ph.D, candidate in the School of Information and Communication Engineering, Dalian University of Technology (DUT). Her research interest is in person re-identification.
\end{IEEEbiography}
\begin{IEEEbiography}{\textbf{Pingping Zhang}}
received his B.E. degree in mathematics and applied mathematics, Henan Normal University (HNU), Xinxiang, China, in 2012. He is currently a Ph.D. candidate in the School of Information and Communication Engineering, Dalian University of Technology (DUT), Dalian, China. His research interests are in deep learning, saliency detection, object tracking and semantic segmentation.
\end{IEEEbiography}
\begin{IEEEbiography}{\textbf{Huchuan Lu}}
(SM'12) received the M.Sc. degree in signal and information processing, PhD degree in system engineering, Dalian
University of Technology (DUT), China, in 1998 and 2008 respectively. He has been a faculty since 1998 and a professor since 2012 in the School of Information and Communication Engineering of DUT. His research interests are in the areas of computer vision and pattern recognition. In recent years, he focus on visual tracking, saliency detection and semantic segmentation. Now, he serves as an associate editor of the IEEE Transactions On Systems, Man, and Cybernetics: Part B.
\end{IEEEbiography}
\begin{IEEEbiography}{\textbf{Hongyu Wang}} 
(M'98) received the B.S. degree from Jilin University of Technology, Changchun, China, in 1990 and the M.S. degree from the Graduate School of Chinese Academy of Sciences, Beijing, China, in 1993, both in electronic engineering. He received the Ph.D. degree in precision instrument and optoelectronics engineering from Tianjin University, Tianjin, China, in 1997. He is currently a Professor with Dalian University of Technology, Dalian, China. His research interests include algorithmic, optimization, and performance issues in wireless ad hoc, mesh, and sensor networks.
\end{IEEEbiography}

\ignore{
\begin{IEEEbiography}{}

\end{IEEEbiography}

\begin{IEEEbiographynophoto}{John Doe}

\end{IEEEbiographynophoto}


\begin{IEEEbiographynophoto}{Jane Doe}
Biography text here.
\end{IEEEbiographynophoto}

}



\end{document}